\documentclass[letterpaper, 10 pt, conference]{ieeeconf}  

\IEEEoverridecommandlockouts                              

\usepackage{mymacros}
\usepackage{booktabs}
\usepackage{tabularx}

\usepackage{color}
\usepackage[dvipsnames]{xcolor}

\makeatletter
\long\def\@makecaption#1#2{%
  \ifx\@captype\@IEEEtablestring%
    \setbox\@tempboxa\hbox{\footnotesize #1.~~ #2}%
    \ifdim \wd\@tempboxa >\hsize%
      \setbox\@tempboxa\hbox{\footnotesize #1.~~ }%
      \parbox[t]{\hsize}{\footnotesize \noindent\unhbox\@tempboxa#2}%
    \else%
      \hbox to\hsize{\footnotesize\hfil\box\@tempboxa\hfil}%
    \fi
    \@IEEEtablecaptionsepspace
  \else
    \@IEEEfigurecaptionsepspace
    \setbox\@tempboxa\hbox{\footnotesize #1.~~ #2}%
    \ifdim \wd\@tempboxa >\hsize%
      \setbox\@tempboxa\hbox{\footnotesize #1.~~ }%
      \parbox[t]{\hsize}{\footnotesize \noindent\unhbox\@tempboxa#2}%
    \else%
      \ifcenterfigcaptions \hbox to\hsize{\footnotesize\hfil\box\@tempboxa\hfil}%
      \else \hbox to\hsize{\footnotesize\box\@tempboxa\hfil}%
      \fi\fi
  \fi}
\makeatother

\title{\LARGE \bf
LILAC: Language‑Conditioned Object‑Centric Optical Flow for Open‑Loop Trajectory Generation
}

\author{Motonari Kambara$^{1}$, Koki Seno$^{1}$, Tomoya Kaichi$^{2}$, Yanan Wang$^{2}$, Komei Sugiura$^{1}$ 
\thanks{$^{1}$Keio University, $^{2}$KDDI Research Inc. motonari.k714@keio.jp}
\thanks{This work was partially supported by JSPS KAKENHI Grant Number 23K28168, JST Moonshot, and JSPS Fellows Grant Number JP23KJ1917.}%
}

\begin{document}

\makeatletter
\let\@oldmaketitle\@maketitle 
\renewcommand{\@maketitle}{\@oldmaketitle 
    \vspace{-4mm}
    \centering

}
\makeatother
\newcommand{\shiftDocumentDownWithMargins}[1]{%
  \addtolength{\topmargin}{#1}      
  \addtolength{\textheight}{-#1}   
  \addtolength{\footskip}{#1}      
  \addtolength{\textheight}{#1}    
}
\maketitle
\vspace{-2mm}
\thispagestyle{empty}
\pagestyle{empty}
\shiftDocumentDownWithMargins{3mm}

\begin{abstract}

We address language-conditioned robotic manipulation using flow-based trajectory generation, which enables training on human and web videos of object manipulation and requires only minimal embodiment-specific data.
This task is challenging, as object trajectory generation from pre-manipulation images and natural language instructions requires appropriate instruction-flow alignment.
To tackle this challenge, we propose the flow-based
Language Instruction-guided open-Loop ACtion generator (LILAC).
This flow-based Vision-Language-Action model (VLA) generates object-centric 2D optical flow from an RGB image and a natural language instruction, and converts the flow into a 6-DoF manipulator trajectory. 
LILAC incorporates two key components: Semantic Alignment Loss, which strengthens language conditioning to generate instruction-aligned optical flow, and Prompt-Conditioned Cross-Modal Adapter, which aligns learned visual prompts with image and text features to provide rich cues for flow generation.
Experimentally, our method outperformed existing approaches in generated flow quality across multiple benchmarks.
Furthermore, in physical object manipulation experiments using free-form instructions, LILAC demonstrated a superior task success rate compared to existing methods.
The project page is available at \url{https://lilac-75srg.kinsta.page/}.
\end{abstract}
\section{Introduction} \label{intro}
Object manipulation is a foundational capability of service robots operating in human environments, requiring safety, efficiency, and task-level generality.
The emergence of increasingly large multimodal datasets~\cite{khazatsky2024droid,ebert2021bridge,walke2023bridgedata} has driven the development of Vision-Language-Action models (VLAs) for robotic manipulation~\cite{brohan2022rt,o2024open,kim2024openvla,black2024pi_0,octo_2023,li2024cogact,wen2025tinyvla,liao2024llara}.
These models are particularly important for service robots, where the ability to appropriately understand and execute human instructions is required~\cite{korekata2025dm2rm, goko2024task}.
Although recent VLAs execute actions contained in their training data with high fidelity, their zero-shot transfer performance remains limited~\cite{black2024pi_0}.  

We address this limitation by recasting manipulation planning as a flow-based prediction problem. 
Given a single RGB image before manipulation and a natural language instruction, this approach jointly predicts (i) an object-centric 2D optical-flow field and (ii) its corresponding manipulator trajectory.  
This decomposition provides a dense supervisory signal that is independent of specific embodiments and enhances generalization to unseen objects, poses, and instructions.
Encoding motion in pixel space provides dense supervision, allows training on the vast supply of web videos that depict human object handling, and reduces the amount of costly embodiment-specific data required for adaptation to new tasks.

Despite these advantages, existing multimodal flow-based approaches~\cite{bharadhwaj2024track2act, gao2024flip, xuflow} have two fundamental limitations.
First, most VLAs are trained on closed-domain datasets that contain narrowly scoped, task-specific data.
As a result, when encountering out-of-distribution scenes, they often generate motions that appear visually plausible but do not align with the given instructions.
Second, most multimodal flow-based approaches~\cite{bharadhwaj2024track2act, gao2024flip, xuflow} use a closed-loop (online) trajectory generation scheme.
At each step, they re-encode the language instruction and current image to predict the next flow or waypoint.
\textcolor{black}{ This approach increases inference cost because the model repeatedly encodes the instruction and current image at each step. 
It can also lead to error accumulation over execution, and it requires expert data that covers many intermediate states.}

\begin{figure}[t]
    \centering
    \includegraphics[width=\linewidth]{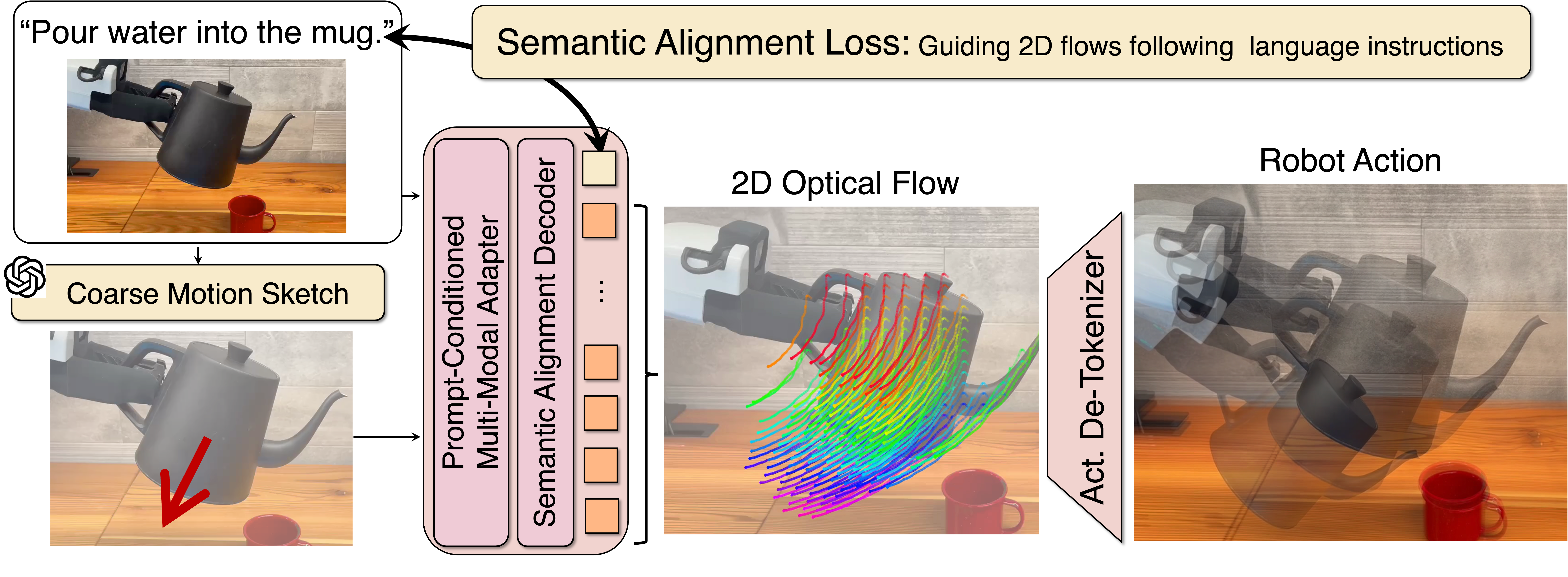}
    \caption{
    Overview of LILAC, 2D object-centric optical flow-based Vision-and-Language trajectory generation framework. In this figure, `Act. De-Tokenizer' represents Action De-Tokenizer. Given a natural language instruction, LILAC generates 2D flow from an RGB image and the instruction, and converts the flow into a 6-DoF robot trajectory.
    }
    \label{fig:eye-catch}
    \vspace{-20pt}
\end{figure}
To overcome these constraints, we develop a flow-based Language Instruction-guided open-Loop ACtion generator (LILAC) capable of generating the entire manipulation trajectory offline, in a single forward pass. 
Fig.~\ref{fig:eye-catch} shows an overview of the proposed method.
The main difference between LILAC and existing methods lies in the introduction of visual prompts and Semantic Alignment Loss.
Unlike existing methods that directly generate 2D flow from images and instructions, LILAC uses visual prompts as intermediate representations.
This enables gradual transformation into dense motion representations, leading to more efficient learning.
Furthermore, Semantic Alignment Loss prevents a common failure mode where models over-rely on visual information and ignore language instructions.
This ensures that generated flows remain consistent with the given commands.
The key contributions of this paper are as follows:
\begin{itemize}
    \item We propose LILAC, a vision-and-language open-loop trajectory generation framework that generates object-centric 2D optical flow from RGB images and natural language instructions, then converts the flow into 6-DoF manipulator trajectories.
    \item We introduce Prompt-Conditioned Multimodal Adapter that integrates images, language instructions, and visual prompts to enable task-adaptive flow generation.
    \item We introduce Semantic Alignment Loss that explicitly encourages the model to learn semantically meaningful representations of language instructions, improving alignment between instructions and generated 2D flow.
\end{itemize}
\section{Related Work \label{sec2}}

\subsection{Vision-Language-Action Models} 
VLAs have been widely studied~\cite{wen2025tinyvla, liao2024llara, lin2025evo, black2024pi_0, 2025pi05}.
Some research suggests that vision-language models can process natural language instructions with visual observations to directly generate robot trajectories~\cite{Driess2023palme, huang2023voxposer, goyal2025vla, wang2025vla}. 
These models implicitly learn a shared representation that aligns language instructions with visual information and encodes embodiment-specific priors for physical execution.
These approaches are highly relevant to our method in that they also execute tasks conditioned on language instructions.

Existing VLA models~\cite{black2024pi_0, 2025pi05, o2024open} face the challenge of requiring a large mount of embodiment-specific data collected using real robots and environments for training, resulting in extremely high training costs.
For instance, $\pi_0$~\cite{black2024pi_0} is pre-trained with approximately 10,000 hours of demonstrations.
Notably, \cite{o2024open} shows that models trained with larger datasets consistently outperform those trained with smaller datasets.
Despite being pre-trained on large amounts of data, these models often exhibit insufficient zero-shot performance.
For example, OpenVLA~\cite{kim2024openvla}, a state-of-the-art VLA, has a success rate of almost 0\% in zero-shot manipulation tasks~\cite{black2024pi_0}.
In contrast, 
the flow generator can be trained with both robot and human manipulation videos, allowing the use of any large-scale dataset that involves human object interactions.
This enables the use of a wide range of video datasets, including videos of object manipulations by both robotic manipulators and humans, offering the advantage of collecting large-scale training data at low cost.     

\subsection{Flow-based Action Generation Models}
Object-centric flow is a promising approach to represent atomic action effectively. 
Indeed, many flow-based action generation models have been developed and reported, representing a wide variety of object manipulations~\cite{bharadhwaj2024track2act, gao2024flip, xuflow, wen2023any}. 
A notable advantage of flow-based action generation models is their requirement for small robot demonstration data. 
In some methods, a small amount of demonstration data is required for trajectory generation because the manipulator trajectory can be obtained simply by accounting for flow discrepancy caused by end-effector embodiment~\cite{bharadhwaj2024track2act}.
Moreover, most existing flow-based models do not require conditioning of natural language instructions for object manipulation. 
Instead, they rely on images of pre- and post-manipulation states as conditions, and they use images taken during or after object manipulation as input.
However, in practice, it is important to be able to execute task instructions given in natural language. 
For this reason, LILAC handles object-manipulation instruction sentences as input.
\section{Problem Settings}
We address language instruction-guided object manipulation using flow-based trajectory generation.
This task consists of two subtasks: 2D object-centric flow generation and manipulator trajectory generation.
In the 2D object-centric flow generation task, given an input RGB image $\bm{x}_{\mathrm{img}}$ and natural language instruction $\bm{x}_{\mathrm{inst}}$, models should generate 2D optical flow $\mathcal{T}_p = \{(x_p^{(t)}, y_p^{(t)})\}_{t=0}^{H-1}$ representing the expected object motion, where $H$ denotes the sequence length. 
We assume that $\{(x_p^{(0)}, y_p^{(0)})\}$ is given.
These flows collectively represent the required motion pattern to execute the task specified by $\bm{x}_{\mathrm{inst}}$ within the visual context $\bm{x}_{\mathrm{img}}$.
Subsequently, in the trajectory generation task, the model is required to generate the manipulator trajectory from $\bm{x}_{\mathrm{img}}$, $\bm{x}_{\mathrm{inst}}$, $\mathcal{T}_p$, the depth image, and camera parameters.


Not using depth images for 2D flow generation is effective in terms of scaling up datasets. 
Standard Large-scale video datasets based on web videos often do not include depth images~\cite{zhou2018towards, miech2019howto100m, Damen2021PAMI, goyal2017something}. 
In 2D flow generation tasks, it becomes possible to use these datasets directly when depth images are not included in the input.
As a result, the model can be scaled up using datasets that consist only of RGB images and language instructions.

\section{Proposed Method}

\begin{figure*}[t]
  \centering
  \includegraphics[width=\textwidth]{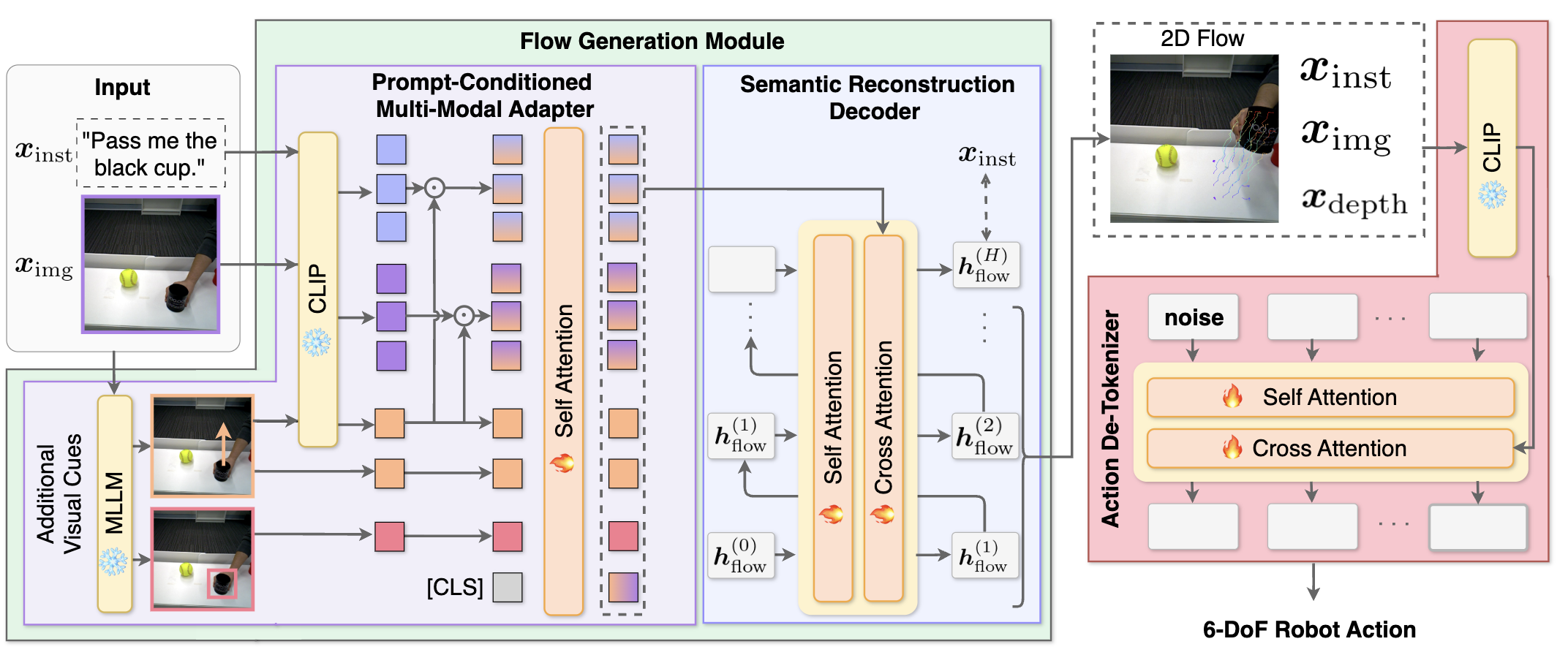}
  \caption{Overview of the LILAC framework.  
  The inputs to the flow generation module are a single RGB image $\bm{x}_{\mathrm{img}}$ and a natural language instruction $\bm{x}_{\mathrm{inst}}$. 
  In this module, first, visual prompts and bounding boxes are generated by the MLLM. Subsequently, 2D flow is autoregressively generated by the Prompt-Conditioned Multi-Modal Adapter and the Semantic Reconstruction Decoder. 
  After generating the 2D flow, based on the flow, $\bm{x}_{\mathrm{img}}$, $\bm{x}_{\mathrm{inst}}$, and $\bm{x}_{\mathrm{depth}}$, a 6-DoF manipulator trajectory is generated by the Action De-Tokenizer.
  }
  \label{fig:overview}
  \vspace{-5mm}
\end{figure*}

We develop LILAC, a novel framework for generating task-relevant motion trajectories from visual observations and natural-language instructions.
Fig.~\ref{fig:overview} shows the network overview. 
LILAC consists of two main modules: a flow generation module and an action de-tokenizer.

\subsection{Flow Generation Module}
\subsubsection{\textbf{Prompt-Conditioned Multi-Modal Adapter}}
We introduce Prompt-Conditioned Multi-Modal Adapter to effectively align the language instruction with the visual scene and to adapt to varying task requirements. 
Existing approaches attempt to align natural language instructions with the resulting object motion (optical flow) directly. 
However, aligning these two modalities directly is challenging because natural language instructions provide high-level semantic information, while optical flow represents low-level, dense motion patterns. 
Our module narrows this granularity gap by first distilling a coarse motion sketch from the instruction and injecting this sketch into the encoder, in the form of a visual prompt. 
The prompt guides the network to acquire a flow-centric latent representation that is aligned with the high-level language description, thus increasing the reliability of language motion correspondence.
The inputs for this module are $\bm{x}_{\mathrm{img}}$ and $\bm{x}_{\mathrm{inst}}$.

\begin{figure}[t]
    \centering
    \includegraphics[width=70mm]{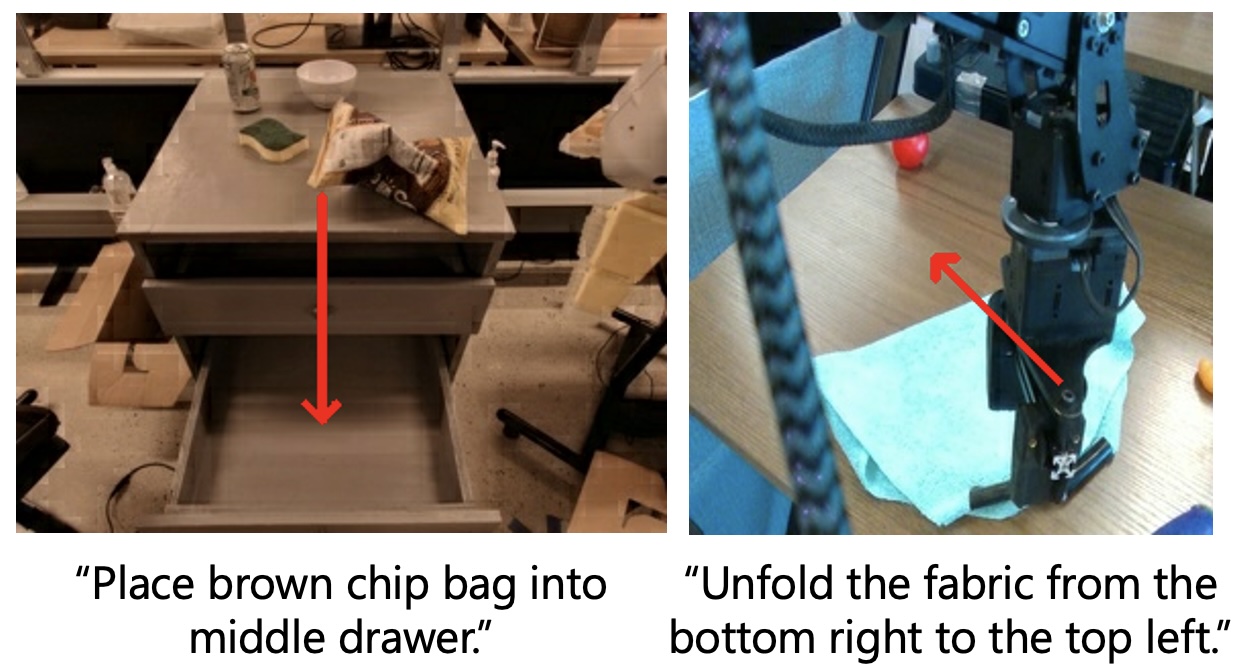}
    \caption{Examples of visual prompts generated by GPT-4o. GPT-4o outputs the start and end points of an arrow that represents the rough shape of the action from a text prompt and an RGB image as input. In this figure, the red arrows indicate the rendered visual prompts.}
    \label{fig:visual_prompt_example}
    \vspace{-5mm}
\end{figure}

From the inputs $\bm{x}_{\mathrm{img}}$ and $\bm{x}_{\mathrm{inst}}$, a multimodal large language model (MLLM) is first employed to generate (i) a displacement vector and (ii) a bounding box of the target object for manipulation.
In LILAC, MLLM generates a visual prompt based on $\bm{x}_{\mathrm{txt}}$ and $\bm{x}_{\mathrm{img}}$ to provide auxiliary information during 2D optical flow generation.
We aim to generate a visual prompt in the form of arrows drawn on the image because this approach has been reported to be effective in existing work~\cite{hwang2025motif, kuang2024ram}.
However, direct image inpainting with an MLLM is still challenging.
Therefore, we instruct the MLLM to output the start and end points of the desired arrow, which is subsequently rendered onto the image to create the visual prompt.

Fig.~\ref{fig:visual_prompt_example} shows examples of generated visual prompts.
In this figure, the red arrows indicate the rendered visual prompts.
As illustrated in the figure, the MLLM successfully comprehend the appropriate direction and distance that the target object should move on the basis of $\bm{x}_{\mathrm{txt}}$ and $\bm{x}_{\mathrm{img}}$, and it is able to output this understanding in the form of an arrow.
Ablation studies examining whether the model effectively utilizes the information from the visual prompt to generate appropriate flow are shown in Table~\ref{tab:flow_results}.
The displacement vector is rendered as an arrow and overlaid on the RGB frame, producing an image with a visual prompt $\bm{x}_{\mathrm{vp}}$ as shown in Fig.~\ref{fig:visual_prompt_example}.  

We integrate the three modalities as follows:
\begin{equation}
  \bm{h}_{\mathrm{mm}} \;=\;
  f_{\mathrm{cma}}\bigl(
      \bm{x}_{\mathrm{img}},\;
      \bm{x}_{\mathrm{inst}},\;
      \bm{x}_{\mathrm{vp}}
  \bigr),
  \label{eq:mm_fusion}
\end{equation}
where $f_{\mathrm{cma}}(\cdot)$ represents a cross-modal adapter based on a transformer encoder architecture .  
The multimodal feature $\bm{h}_{\mathrm{mm}}$ serves as a latent flow representation that conditions the subsequent module responsible for predicting the dense optical-flow field.

\subsubsection{\textbf{Semantic Alignment Decoder}}
Semantic Alignment Decoder employs a transformer decoder-based architecture. 
This decoder autoregressively generates 2D flow conditioned on $\bm{h}_{\mathrm{mm}}$.
Given the embeddings of the tracking point sets up to time $t-1$, $[\bm{h}_\mathrm{flow}^{(i)}|i=0,1,\ldots,t-1]$, the embedding of the tracking point set at time $t$, $\bm{h}_\mathrm{flow}^{(t)}$ is obtained as follows:
\begin{align}
\bm{h}_\mathrm{flow}^{(t)} &= f_{ca}\left(f_{sa}\big([\bm{h}_\mathrm{flow}^{(i)}|i=0,1,\ldots,t-1]\big), \bm{h}_{\mathrm{mm}}\right),
\end{align}
where $f_{sa}(\cdot)$ and $f_{ca}(\cdot, \cdot)$ represent self attention and cross attention, respectively.

In the transformer decoder, the output tokens $[\bm{h}_\mathrm{flow}^{(i)}|i=0,1,\ldots,H-1]$ are considered as predicted 2D flow coordinates $\tilde{\mathcal{T}}_p$, whereas $\bm{h}_\mathrm{flow}^{(H)}$ is handled as a CLS token, which is encouraged to align with the language feature representation.
LILAC is trained with a composite loss function designed to encourage appropriate trajectory generation and effective language understanding. 
The objective combines a weighted coordinate prediction loss with our proposed Semantic Alignment Loss $\mathcal{L}_{\text{sem}}$.
This $\mathcal{L}_{\text{sem}}$ encourages the model to preserve semantic information from the instruction throughout the generation process, enhancing the alignment between the command and the generated motion.

%
\begin{table*}[t]
  \centering
  \caption{Quantitative comparison on the Fractal and BridgeData V2 subsets in the Robot Flow benchmark. 
  In addition, the rows ``LILAC w/o srl'' and ``LILAC w/o vp'' respectively show the results of the ablation studies on Semantic Reconstruction Loss and the visual prompt, respectively.
  Lower values of ADE are better. Meanwhile, higher values of AUC and P@K are better. Best values per column are shown in \textbf{bold}.}
  \setlength{\tabcolsep}{4pt}
  \begin{tabular}{lccccccccccc}
    \toprule
    \multirow{2}{*}{Method} & \multicolumn{5}{c}{\textbf{Fractal}} & \multicolumn{5}{c}{\textbf{BridgeData V2}}\\
    \cmidrule(lr){2-6}\cmidrule(lr){7-11}
      & ADE$\downarrow$ & AUC$\uparrow$ & P@5$\uparrow$ [\%] & P@10$\uparrow$ [\%] & P@20$\uparrow$ [\%]
      & ADE$\downarrow$ & AUC$\uparrow$ & P@5$\uparrow$ [\%] & P@10$\uparrow$ [\%] & P@20$\uparrow$ [\%] \\
    \midrule
    Im2Flow2Act~\cite{xuflow}      & 56.46 & 0.190 & 13.4 & 16.2 & 27.4 & 63.19 & 0.177 & 13.3 & 15.4 & 24.5 \\
    FLIP~\cite{gao2024flip}        & 44.41 & 0.165 & 6.9 & 14.1 & 28.4 & 41.95 & 0.200 & 9.8 & 18.0 & 32.2 \\
    LILAC w/o srl                  & 28.94 & 0.357 & 14.8 & 32.0 & 60.2 & 31.14 & 0.325 & 16.3 & 30.0 & 51.2 \\
    LILAC w/o vp                   & 33.65 & 0.253 & 9.0 & 20.9 & 45.8 & 37.09 & 0.211 & 9.7 & 17.9 & 35.6  \\
    \textbf{LILAC (full)}    
                                   & \textbf{26.98} & \textbf{0.434} & \textbf{21.4} & \textbf{42.1} & \textbf{66.8}
                                   & \textbf{29.44} & \textbf{0.396} & \textbf{22.9} & \textbf{38.2} & \textbf{57.6} \\
    \bottomrule
  \end{tabular}
  \label{tab:flow_results}
  \vspace{-5mm}
\end{table*}

\begin{figure*}[t]
    \centering
    \includegraphics[width=160mm]{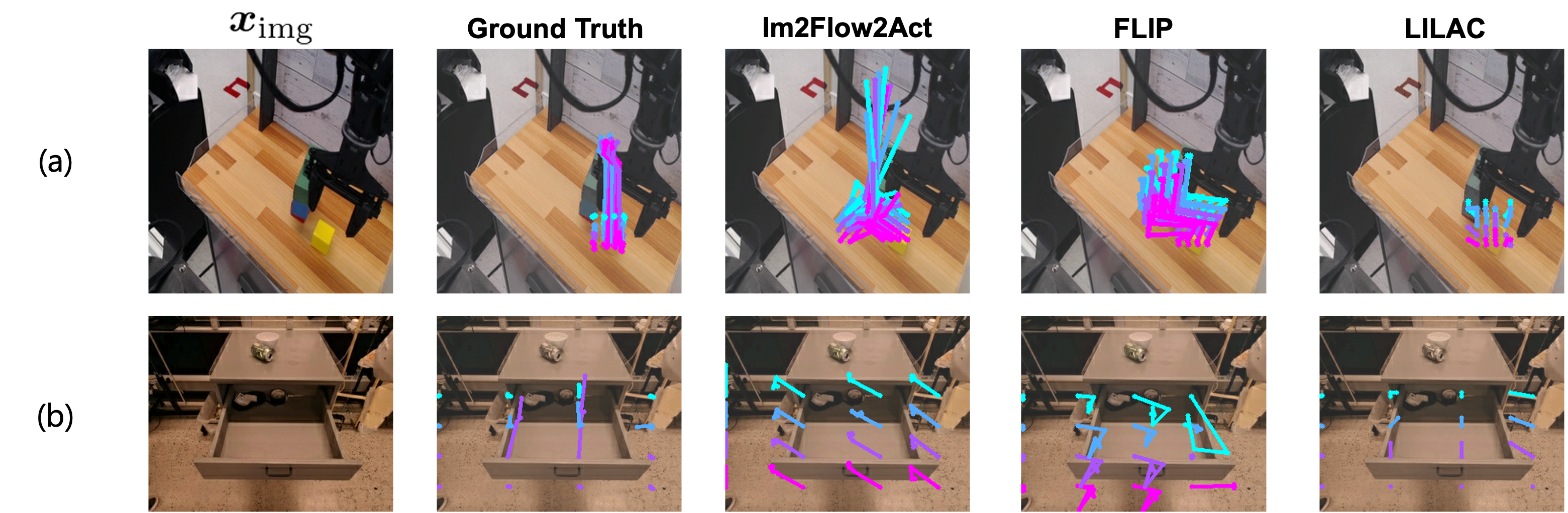}
        \vspace{-2mm}
    \caption{
Qualitative results on the Robot Flow benchmark. The given $\bm{x}_{\mathrm{inst}}$ was (a) ``Take the yellow block to put it on top of the tower.'' and (b) ``Close top drawer.'', respectively. 
    }
    \label{fig:qualitative}
    \vspace{-5mm}
\end{figure*}

\subsection{Action De-Tokenizer}
The conversion process is inspired by Track2Act~\cite{bharadhwaj2024track2act}.
The inputs to this module are the 2D flow sequence, depth image $\bm{x}_{\mathrm{depth}}$, and camera parameters.
The output is a sequence of end-effector trajectories.

The module consists of two stages: a first stage generating a coarse trajectory and a second stage generating a refined trajectory.
In the first stage, a coarse 6-DoF manipulator trajectory is estimated via rigid transformation estimation, using the predicted flow, $\bm{x}_{\mathrm{depth}}$, and camera parameters.
This stage employs the algorithm used in \cite{bharadhwaj2024track2act}.
In the second stage, the generated trajectory is refined using a transformer decoder.
The input to the transformer decoder is a token sequence formed by concatenating the generated coarse trajectory and the predicted flow along the sequence dimension. This process is conditioned on $\bm{x}_{\mathrm{txt}}$ and $\bm{x}_{\mathrm{img}}$.
The output is the refined 6-DoF manipulator trajectory.
At each time step within the decoder, self-attention is computed with input sequence tokens, followed by cross-attention conditioning on $\bm{x}_{\mathrm{txt}}$ and $\bm{x}_{\mathrm{img}}$.
Unlike Track2Act, we employ an auto-regressive model instead of a diffusion-based model, primarily to increase the convergence speed.

\subsection{Loss Function}
We use two loss functions for LILAC.
The first loss function $\mathcal{L}_{\mathrm{fg}}$ is used in the flow-generation module:
\begin{equation}
\mathcal{L}_{\mathrm{fg}}
=\mathcal{L}_{\text{sem}}(\bm{h}_{\mathrm{inst}}, \bm{h}_\mathrm{flow}^{(H)})
+\mathcal{L}_{\text{flow}}(\mathcal{T}_p, p(\tilde{\mathcal{T}}_p)),
\end{equation}
where $\mathcal{L}_{\text{flow}}(\cdot)$, $\bm{h}_{\mathrm{inst}}$, and $p(\tilde{\mathcal{T}}_p)$ represent a cross-entropy loss, the embeddings of $\bm{x}_{\mathrm{inst}}$ with the CLIP language encoder, and the sequence of predicted probability for each coordinate of the 2D flow, respectively.
Moreover, $\mathcal{L}_{\text{sem}}$ is a loss that encourages the 2D flow representation to align with the instruction.
Specifically, it aligns the final token of the 2D flow sequence with $\bm{h}_{\mathrm{inst}}$, ensuring that the 2D flow encodes a motion representation conditioned on the language instruction.
We employ an L1 loss for this objective.
For $\mathcal{L}_{\text{flow}}$, we adopt a cross-entropy loss based on the following reason.
Given an input image of size $W\times H$, the horizontal and vertical flow components are cast as classification problems of independent $W$ class and $H$ class, respectively. 
Compared with direct regression, framing flow estimation as classification can reduce the solution space, allowing the network to converge with fewer training samples and iterations.

The second loss function is used in Action De-Tokenizer.
This module is trained with an L1 loss, which measures the distance between the teleoperated end-effector trajectory and the output refined trajectory.
\section{Experiments}
\subsection{Experimental Settings \label{ex/flow}}
We constructed the Robot Flow benchmark to provide object-centric optical flow annotations for manipulation tasks.
Our benchmark is built on the Fractal~\cite{brohan2022rt} and BridgeData V2~\cite{walke2023bridgedata} datasets.
We describe the specific construction procedure below.
Firstly, we used Qwen-2.5-VL~\cite{ahmed2025qwen} to generate the bounding box of the target object based on a natural language instruction.
Next, we uniformly sampled the $n$ tracking points within the generated bounding box.
We jointly tracked those points throughout the whole video using CoTracker-3~\cite{karaev2024cotracker3}.
The resulting trajectories were converted to optical flow and used as ground-truth. 
Following the way in \cite{bharadhwaj2024track2act}, we sampled $H$ steps from the flows. 
In the experiments, we set $H=8$.

The Robot Flow benchmark consists of two subsets: the Fractal subset and the BridgeData V2 subset.
The Fractal subset comprises a total of 26,516 episodes, with 23,866 episodes for training, 1,325 episodes for validation, and 1,325 episodes for testing.
The BridgeData V2 subset includes 10,672 episodes, divided into 9,605, 533, and 534 for training, validation, and testing, respectively.
During training, we used the combined training sets from both subsets.

For visual prompt generation in the flow generation module, we compared with both GPT-4o~\cite{gpt-4o}, a representative MLLM, and Qwen-2.5-VL.
We found that the prompts generated by GPT-4o were qualitatively superior to those generated by Qwen-2.5-VL. 
Therefore, we adopted GPT-4o for visual prompt generation.

In the experiments of 2D flow generation, we used two baseline methods: Im2Flow2Act~\cite{xuflow} and FLIP.
Im2Flow2Act and FLIP were selected as representative examples of flow-based VLAs.
Training was carried out using an NVIDIA H200 GPU. 
\begin{figure}[t]
    \centering
    \includegraphics[width=\linewidth]{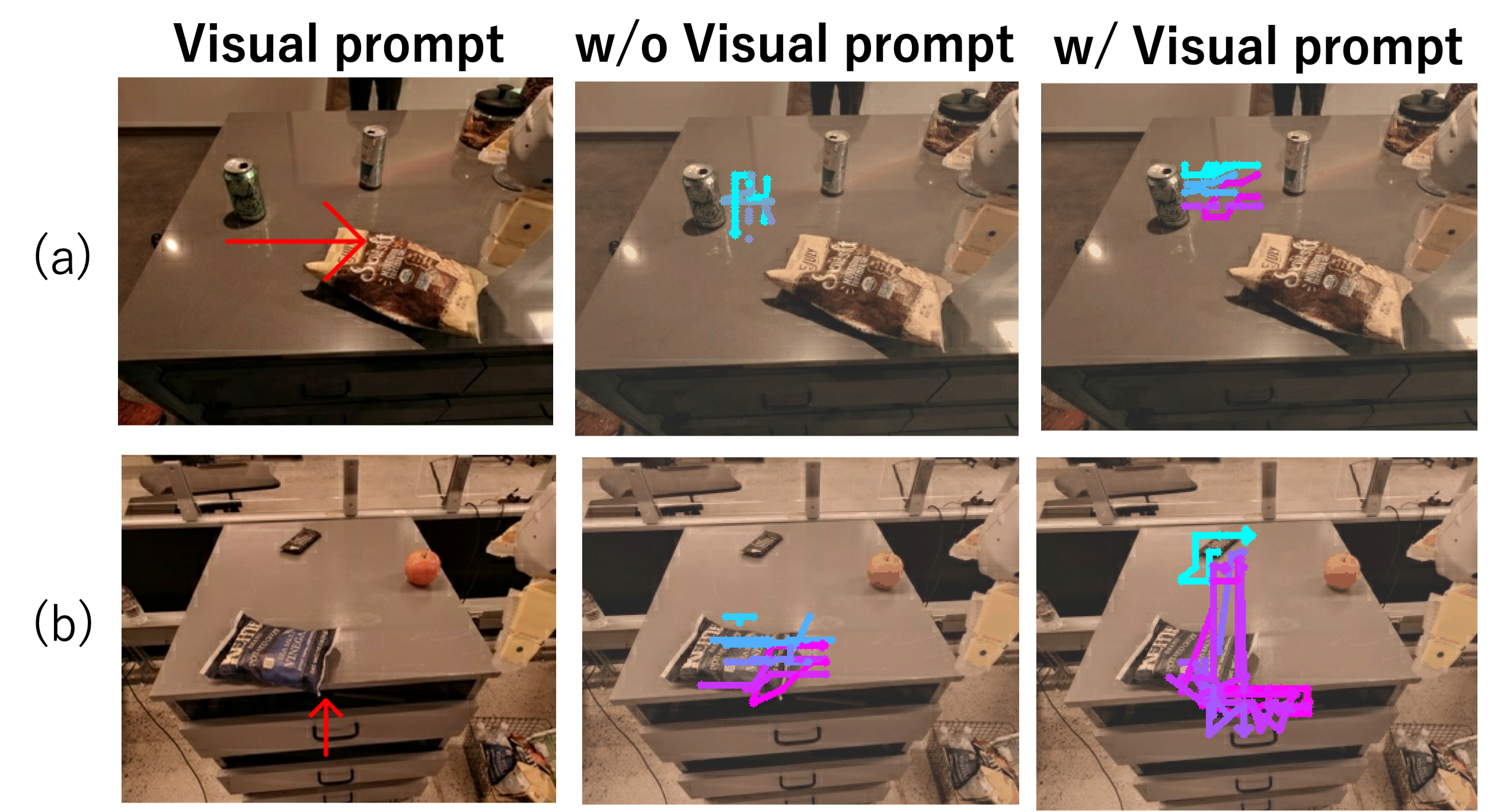}
    \caption{\textcolor{black}{ Qualitative results with and without visual prompts. `Visual prompt', `w/o Visual prompt', and `w/ Visual prompt' denote the visual prompt generated by the MLLM, the flow generated without a visual prompt, and the flow generated with a visual prompt, respectively. In each example, the given instruction sentence is as follows: (a) ``Move 7up can near brown chip bag.'' (b) ``Pick black chip bag.''}}
    \label{fig:vp_qual}
    \vspace{-5mm}
\end{figure}

\subsection{Quantitative Results}
Table~\ref{tab:flow_results} shows the quantitative comparison results of LILAC and the baseline methods on the Robot Flow benchmark. 
The best value in each column is presented in bold. 
We adopted the standard metrics used in prior works~\cite{bharadhwaj2024track2act, gao2024flip}: Average Distance Error (ADE), Precision@K (P@K), and the area under the curve (AUC). 
Here, P@K denotes the proportion of predicted points falling within a Euclidean distance \(K\) of the ground-truth waypoint. 
In the experiments, we set K to 5, 10, and 20.
Following the protocols of~\cite{gao2024flip, bharadhwaj2024track2act}, we evaluated only points whose displacement from the previous timestep exceeded a threshold $\delta_t$.

As shown in the table, LILAC outperformed all baseline methods across all metrics.
Compared to the strongest baseline, LILAC reduced ADE by 17.43 points and improved AUC by 0.244 on the Fractal subset.
On the BridgeData V2 subset, it reduced ADE by 12.51 points while achieving the highest AUC score.
These results demonstrate that LILAC generates more accurate 2D flow than baseline methods across multiple domains.
\subsection{Qualitative Results}
Fig.~\ref{fig:qualitative} shows the qualitative results of LILAC and the baseline methods. 
For example, in Fig.~\ref{fig:qualitative}(a), the manipulator was instructed to place the yellow block on top of the tower. 
The baseline methods failed to generate flows that moved the block in a straight path toward the tower. 
In contrast, the flow generated by our method appropriately indicates the direction in which the block should be moved.
%
%
These examples demonstrate that the proposed method is capable of generating 2D flow that is properly aligned with the given natural language instructions and visual inputs.

\textcolor{black}{
We also present qualitative examples of flows generated with and without visual prompts in Fig.~\ref{fig:vp_qual}. 
For example, in Fig.~\ref{fig:vp_qual}(a), when no visual prompt was used, the generated flow was directed upward, whereas when a visual prompt was used, the flow was appropriately generated horizontally.
These examples show that using appropriate visual prompts could improve the quality of the generated flow in some cases. 
}

\begin{table}[t]
    \centering
    \textcolor{black}{
    \caption{Error categories in the visual prompt failure analysis. We analyzed 100 examples randomly sampled from the Fractal subset, and identified 37 failure cases.}
    \label{tab:vp_analysis}
    \begin{tabular}{lcc}
        \toprule
        Error category & \#Samples\\
        \midrule
        Incorrect end point         & 12 \\
        Overly short visual prompt  & 6  \\
        Incorrect start point       & 5 \\
        Invalid visual prompt       & 5 \\
        Reversed direction          & 5 \\
        No visual prompt generated  & 4 \\
        \midrule
        Total                       & 37 \\
        \bottomrule
    \end{tabular}
    }
        \vspace{-3mm}
\end{table}
\begin{figure}[t]
    \centering
    \includegraphics[width=60mm]{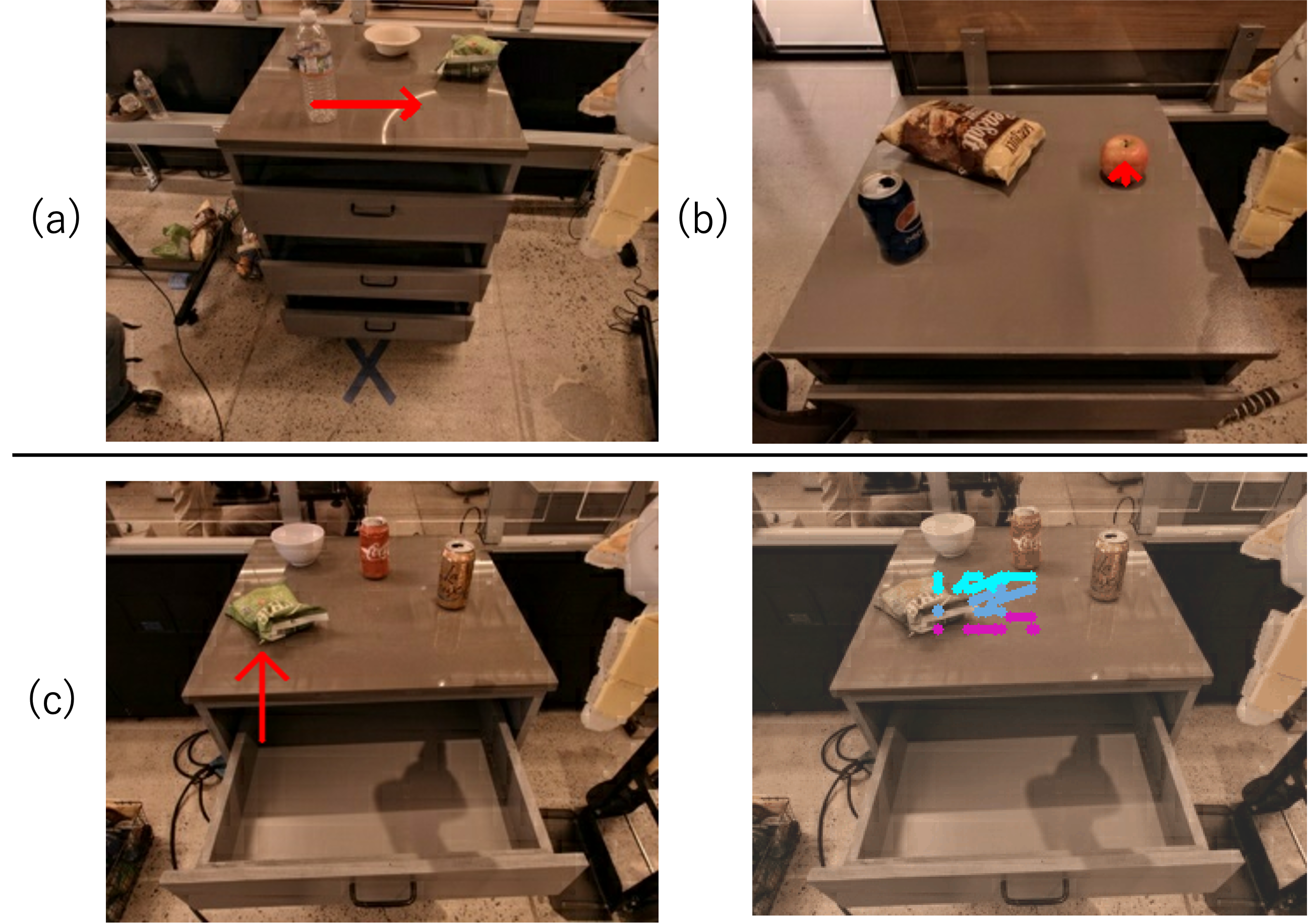}
    \textcolor{black}{
    \caption{Top row: (a) Representative example of Incorrect end point. In this example, the instruction ``Move water bottle near paper bowl.'' was given. 
    (b) Representative example of `Overly short visual prompt'. The given instruction sentence was ``Pick apple.''
    Bottom row: Example of flow generation conditioned on an incorrect visual prompt. The instruction ``Place green jalapeno chip bag into top drawer.'' was given. 
    }
    }
    \label{fig:vp_fail}
    \vspace{-5mm}
\end{figure}

\begin{figure}[t]
    \centering
    \includegraphics[width=70mm]{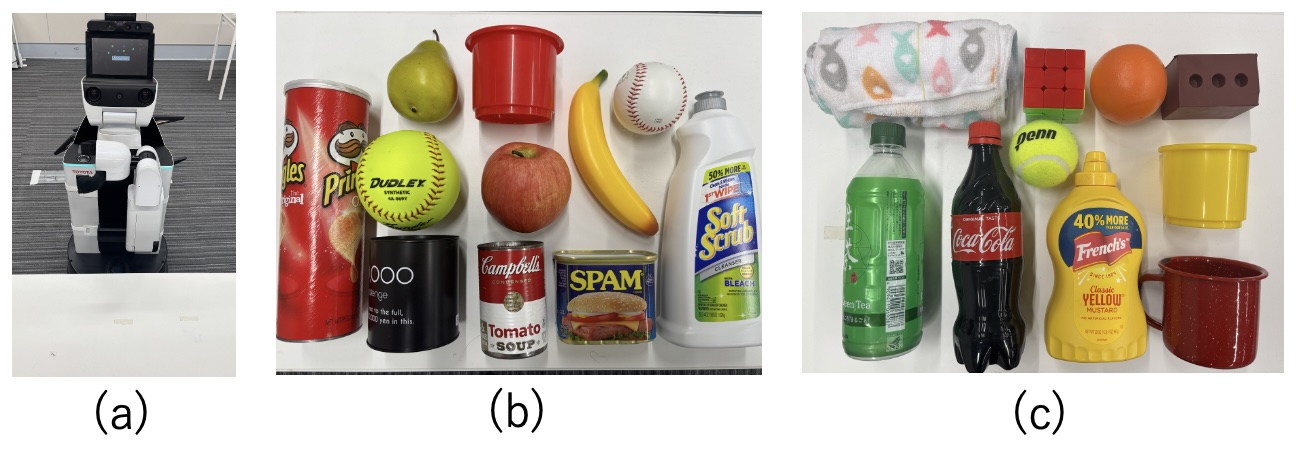}
            \vspace{-2mm}
    \caption{(a) HSR used as the robot platform; (b) objects used for data collection; (c) objects used for evaluation.}
    \label{fig:pe_settings}
    \vspace{-5mm}
\end{figure}

\subsection{Ablation Studies}
To validate the effectiveness of our framework, we conducted two ablation studies.
Specifically, we conducted the ablation studies on the construction of the semantic reconstruction loss (w/o SRL) and visual prompt (w/o VP).
As shown in Table~\ref{tab:flow_results}, model performance declined in both cases, indicating that these novel components contributed to the overall performance improvement. 
In particular, when the visual prompt was removed, the primary metric ADE degraded by 6.67 and 7.65 points on the Fractal and BridgeData V2 subsets, respectively. 
This demonstrates that the visual prompt was the most significant factor contributing to performance improvement.

\textcolor{black}{
\subsection{Error Analysis}
To analyze how visual prompts affect flow generation, we conducted an additional subject experiment. 
Specifically, we randomly sampled 100 outputs generated by LILAC and asked subjects to evaluate the quality of the visual prompts.
As a result, 37 samples were classified as failure cases.
%
%
We further categorized the failures into six error types, and the counts for each category are summarized in Table~\ref{tab:vp_analysis}.
We defined these categories as follows:
(i) \textbf{Incorrect end point}, where the end point of the visual prompt is incorrect as shown in Fig.~\ref{fig:vp_fail}(a);
(ii) \textbf{Overly short visual prompt}, where the prompt is too short to adequately indicate the intended motion as shown in Fig.~\ref{fig:vp_fail}(b);
(iii) \textbf{Incorrect start point}, where the start point of the visual prompt is incorrect;
(iv) \textbf{Invalid visual prompt}, where the generated visual prompt is clearly invalid;
(v) \textbf{No visual prompt generated}, where no visual prompt is output; and
(vi) \textbf{Reversed direction}, where the arrow direction is opposite to the intended direction.
This result shows that endpoint-related errors were the most frequent in our analysis, suggesting that appropriately specifying the target location remains a key challenge.
}

\section{Physical Experiments}
\subsection{Experimental Settings}
\begin{figure*}[t]
    \centering
    \includegraphics[width=175mm]{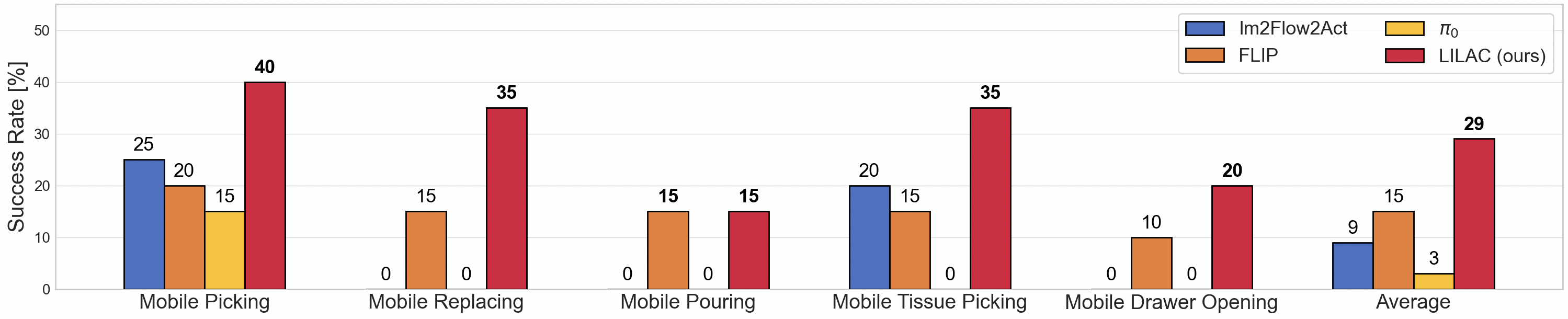}
    \vspace{-10pt}
    \caption{
Success rates of different methods on physical experiments. We compared our proposed method against three baseline approaches (Im2Flow2Act, FLIP, and $\pi_0$) across five manipulation tasks and report the average success rate. Each task was evaluated over 20 trials. \textbf{Bold numbers} indicate the highest success rate for each task. 
    }
    \label{fig:physical_quant}
    \vspace{-5mm}
\end{figure*}
\textcolor{black}{
In the physical experiments, we evaluate zero-shot generalization to new scenes using instructions, objects, and furniture layouts not present in the training data.
}
Fig.~\ref{fig:pe_settings}~illustrates the experimental setup used in the physical experiments showing the robot platform, objects, and environment.
We conducted experiments on five tasks: Mobile Replacing, Mobile Moving, Mobile Pouring, Mobile Tissue Picking, and Mobile Drawer Opening. 
Mobile Replacing, Mobile Moving, and Mobile Pouring are extensions of standard picking, moving, and pouring tasks that additionally require robotic base movement. 
The Mobile Tissue Picking task involves extracting a tissue from a tissue box, while the Mobile Drawer Opening task requires opening a drawer. 
We selected these tasks as they are representative benchmarks commonly used in physical experiments.

We used a variety of common household objects (shown in Fig.~\ref{fig:pe_settings}~(b) and (c)) to ensure diversity in terms of visual appearance and physical size.
We used Human Support Robot (HSR)~\cite{yamamoto2019development}, developed by Toyota Motor Corporation, as the robotic platform.
This mobile manipulator has been used as the standard platform for the RoboCup@Home competition~\cite{robocupathome} since 2017.

During the Mobile Replacing, Mobile Moving, and Mobile Pouring tasks, we used a set of 10 object types selected from the object pool shown in Fig.~\ref{fig:pe_settings} (c) that were explicitly held out and not used during fine-tuning data collection.
The evaluation involved the execution of 20 episodes for each task, resulting in a total of 100 evaluation trials.

First, we collected 100 episodes per task performed by human hands to obtain object-centric optical flow data.  
Subsequently, we collected an additional 20 episodes per task performed by the robot, which provided embodiment-specific data for fine-tuning.  
Both LILAC and the baseline methods were trained on the combined dataset and evaluated on unseen objects.
As baseline methods, we adopted Im2Flow2Act, FLIP, and $\pi_0$~\cite{black2024pi_0}.

In the physical experiments, our goal was to verify that the quality of the generated 2D flow correlates with the task success rate on the real robot.
Therefore, for all flow-based VLA baselines, we used the Action De-Tokenizer from LILAC as the module that converts the 2D flow into a 6-DoF robot trajectory.
Also, in this task, each method generates object-centric flow. 
Therefore, the generated trajectory represents the motion after grasping.
For this reason, following existing research~\cite{bharadhwaj2024track2act}, the trajectory up to grasping the target object was generated by a heuristic method.

\subsection{Quantitative Results}
\begin{figure}[h]
    \centering
    \includegraphics[width=70mm]{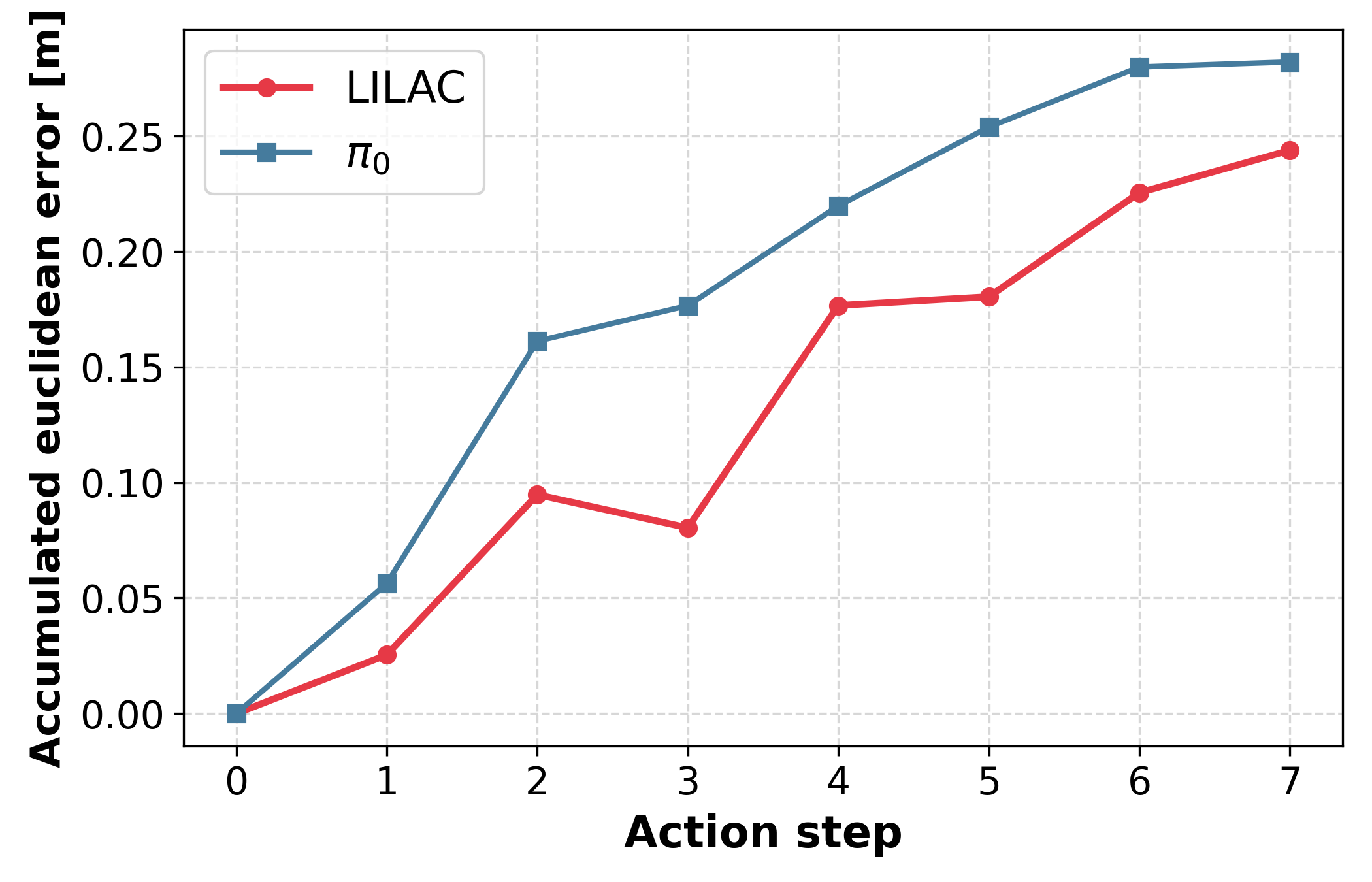}
        \vspace{-2mm}
    \caption{\textcolor{black}{Compounding error at test time. The horizontal axis represents the action step, and the vertical axis represents the Euclidean distance to the human demonstration trajectory. 
    }}
    \label{fig:conpound_error}
    \vspace{-6mm}
\end{figure}

Fig.~\ref{fig:physical_quant} shows the quantitative results. 
The results revealed that LILAC outperformed all baseline methods in each task. 
In terms of the average success rate across all tasks, LILAC outperformed FLIP, the best baseline method, by 14 points.
These findings indicate that the flows predicted by LILAC include more effective action semantics, increasing the higher execution reliability on a real robot.

The primary reason for the low success rate of $\pi_0$ is that object manipulation in our experiments was performed as whole-body manipulation, which involved significant changes in viewpoint. 
Since the pretraining data for $\pi_0$ consisted mostly of table-top manipulation with fixed viewpoints, the pretraining was not effective for our setting.

\textcolor{black}{
Table~\ref{tab:physical_inference_speed} shows quantitative inference speed comparisons. 
As shown in the table, LILAC achieved an inference speed approximately 1.4 times faster than $\pi_0$.
The total inference time gap can become larger for $\pi_0$, which uses closed-loop trajectory generation, than for open-loop trajectory generation methods.
}

\textcolor{black}{
We also conducted a trajectory comparison.
Fig. \ref{fig:conpound_error} shows the result for one randomly selected episode in the Mobile Picking task. 
This result reveals that the trajectories generated by $\pi_0$ exhibit a larger deviation from the reference trajectories than those generated by LILAC.
This observation suggests that LILAC, which uses open-loop trajectory generation, is more robust to compounding errors over time.
}

\subsection{Qualitative Results}
\begin{figure*}[t]
    \centering
    \includegraphics[width=\linewidth]{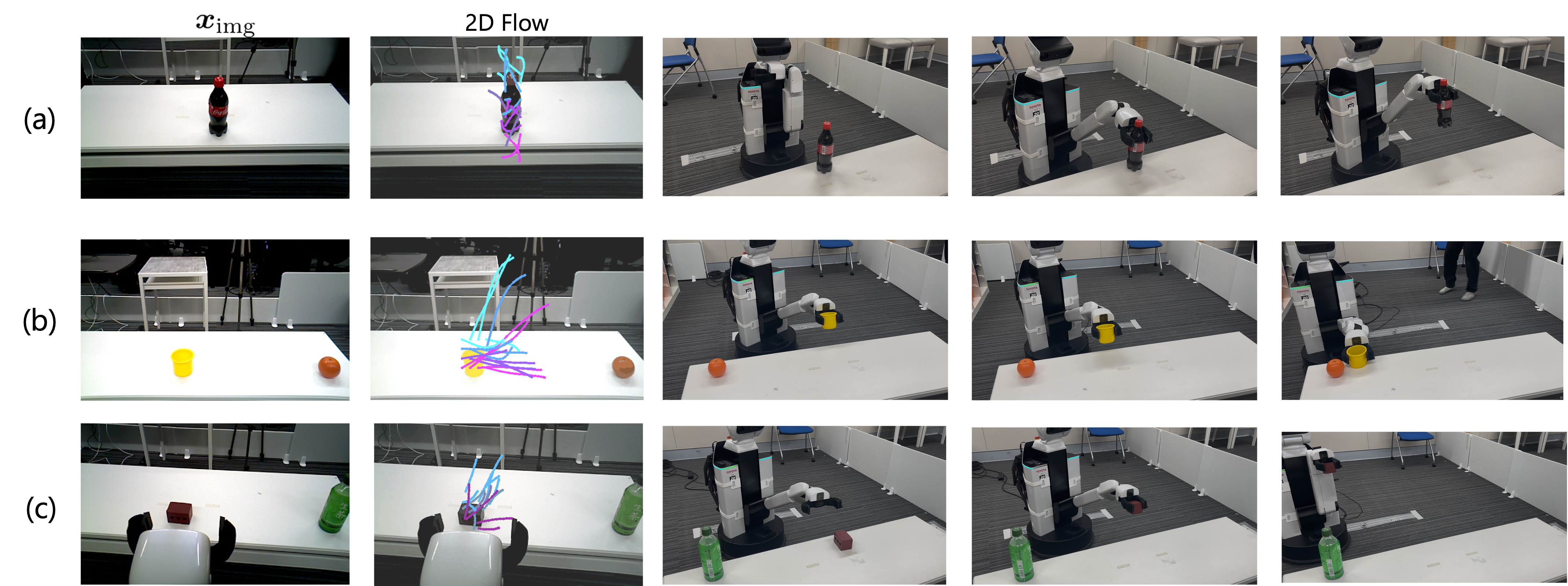}
    \caption{
Qualitative results of LILAC executing tasks on the real HSR platform. In each example,  the following instructions are given: (a) ``Could you take the coke.'' (b) ``Place the cup near the orange.'' (c) ``Put the brick close to the bottle.'' 
In this figure, (a) and (b) represent successful execution examples, and (c) shows a failure case.
The {\bf 2D Flow} column shows the 2D flow generated by the flow generation module.
    }
    \label{fig:real_qual_results}
    \vspace{-6mm}
\end{figure*}
Fig.~\ref{fig:real_qual_results} presents qualitative results for LILAC executing tasks on the real HSR platform.
Fig.~\ref{fig:real_qual_results}~(a) and (b) represent successful execution examples.
In the case shown in Fig.~\ref{fig:real_qual_results}~(a), LILAC generated a flow corresponding to the coke bottle lifted vertically upwards. 
Based on the generated trajectory, the robot successfully grasped and lifted the target bottle.
These cases demonstrate LILAC's capability to generate meaningful end-effector trajectories guided by appropriate flows from its flow generation module, leading to successful task execution.

\begin{table}[t]
    \centering
    \textcolor{black}{
    \caption{Inference speed for trajectory generation in physical experiments. Lower is better. The fastest result is shown in bold.}
    \label{tab:physical_inference_speed}
    \begin{tabular}{lc}
        \toprule
        Method & Inference speed [ms/action] \\
        \midrule
        $\pi_0$~\cite{black2024pi_0}      & 541.0 \\
        Im2Flow2Act~\cite{xuflow}  & 298.7 \\
        FLIP~\cite{gao2024flip}        & \textbf{21.6} \\
        LILAC        & 376.6 \\
        \bottomrule
    \end{tabular}
    }
        \vspace{-8mm}
\end{table}

Fig.~\ref{fig:real_qual_results}~(c) shows a failure case. 
In this episode, the flow generation module generated a flow with a vertical lifting component and some lateral component in the direction of the target bottle. 
As shown in the panels, the robot successfully lifted the brick. 
However, it then unexpectedly moved backward and sideways toward the bottle's location.
Therefore, this episode was considered a failure.
We identified two primary reasons for this suboptimal behavior: (1) The predicted flow was dominated by its vertical component, causing the Action De-tokenizer to generate a trajectory focused primarily on lifting while ignoring the necessary horizontal placement motion; and (2) Converting 2D flow into trajectories, even with the refinement model, likely encountered inherent challenges in accurately inferring depth information.
This limitation could potentially explain the incorrect backward movement observed in the execution.

\section{Conclusion}
In this study, we addressed the language instruction-guided object manipulation using flow-based trajectory generation.
Our main contributions can be summarized as follows:
We proposed LILAC, a VLA framework for open-loop trajectory generation that predicts object-centric 2D optical flow from RGB images and natural language instructions and then converts the predicted flow into manipulator trajectories. 
We introduced a Prompt Conditioned Multimodal Adapter that integrates input information to enable task-adaptive flow generation.
We also introduced Semantic Alignment Loss, which improves alignment between the instructions and the generated 2D flow. 
In addition, LILAC outperformed all baseline methods on both the newly constructed Robot Flow benchmark and in physical robot experiments.

We acknowledge four principal limitations of the present study.
\textbf{(i) Static scene assumption.}
All experiments were conducted in an open-loop setting with a static environment during execution.
This design enables fast inference and may reduce compounding error from recurrent replanning, but the model cannot adapt after trajectory generation if the scene changes unexpectedly.
Dynamic environments would require closed-loop replanning or reactive residual control.
\textbf{(ii) Limited expressiveness of 2D flow.}
Some manipulation skills are difficult to represent with image plane optical flow alone, especially tasks that require depth reasoning or 3D tool motion, such as obstacle-aware placing or stirring.
Because 2D flow has no explicit depth dimension, motions with strong out-of-plane components are hard to encode.
\textcolor{black}{
\textbf{(iii) Sensitivity to visual prompt errors.}
Flow generation can degrade when the MLLM produces an incorrect visual prompt, such as an incorrect start point, target location, invalid prompt, or reversed direction.
This behavior is observed in Fig.~\ref{fig:vp_qual}(c), where both the visual prompt and the generated flow point in the wrong direction, suggesting that prompt errors can propagate to downstream flow generation.
}
\textcolor{black}{
\textbf{(iv) Failure cases in visually and spatially complex scenes.}
MLLM-based visual prompt generation can still fail in scenes with multiple objects and complex spatial relationships.
Although the MLLM succeeds in some multi-object scenes, failures still occur in others (e.g., Fig.~\ref{fig:vp_fail}), indicating that improving visual prompt quality in complex scenes remains an important direction for future work.
}
As future work, the representation could be enriched with depth signals from recent monocular depth estimators, such as Depth Anything~\cite{yang2024depth}.
Incorporating pseudo depth maps into the flow prediction pipeline may enable depth-aware planning while preserving the scalability of RGB language pretraining.




\bibliographystyle{IEEEtran}
\bibliography{reference}

\end{document}